\documentclass[runningheads,a4paper]{llncs}

\usepackage{amssymb}
\usepackage{graphicx}
\usepackage{amsmath}
\usepackage{subcaption}
\usepackage{arabtex}
\usepackage{utf8}
\setcode{utf8}
\usepackage{array}
\usepackage{geometry}
\usepackage{multirow}
\usepackage{longtable}

\usepackage{url}
\urldef{\mailsa}\path|{onajar,akoubaa}@psu.edu.sa|    
\newcommand{\keywords}[1]{\par\addvspace\baselineskip
\noindent\keywordname\enspace\ignorespaces#1}

\begin{document}

\mainmatter  

\title{Enhancing Semantic Similarity Understanding in Arabic NLP with Nested Embedding Learning}

\titlerunning{Nested Arabic Embeddings}

\author{Omer Nacar, Anis Koubaa}
\authorrunning{OMER NACAR}

\institute{Robotics and Internet-of-Things Lab,\\
Prince Sultan University, Riyadh 12435, Saudi Arabia\\
\mailsa\\
\url{http://www.psu.edu.sa}}

\toctitle{Enhancing Semantic Similarity Understanding in Arabic NLP with Nested Embedding Learning}
\tocauthor{OMER NACAR}
\maketitle

\begin{abstract}
This work presents a novel framework for training Arabic nested embedding models through Matryoshka Embedding Learning, leveraging multilingual, Arabic-specific, and English-based models, to highlight the power of nested embeddings models in various Arabic NLP downstream tasks. Our innovative contribution includes the translation of various sentence similarity datasets into Arabic, enabling a comprehensive evaluation framework to compare these models across different dimensions. We trained several nested embedding models on the Arabic Natural Language Inference triplet dataset and assessed their performance using multiple evaluation metrics, including Pearson and Spearman correlations for cosine similarity, Manhattan distance, Euclidean distance, and dot product similarity. The results demonstrate the superior performance of the Matryoshka embedding models, particularly in capturing semantic nuances unique to the Arabic language. Results demonstrated that Arabic Matryoshka embedding models have superior performance in capturing semantic nuances unique to the Arabic language, significantly outperforming traditional models by up to 20-25\% across various similarity metrics. These results underscore the effectiveness of language-specific training and highlight the potential of Matryoshka models in enhancing semantic textual similarity tasks for Arabic NLP.

\end{abstract}

\section{Introduction}

Representation learning \cite{LeCun} forms the backbone of cutting-edge machine learning (ML) systems, offering rich, multidimensional vectors that capture intricate information necessary \cite{Nayak} for various Natural Language Processing (NLP) downstream tasks including  semantic textual similarity, semantic search, paraphrase mining, text classification, clustering, and more. These learned representations are typically static, designed to maintain high-dimensional fidelity across all applications, regardless of their unique resource and accuracy demands. This inherent rigidity often results in inefficiencies, particularly at web-scale \cite{Dean}, where the deployment cost of these embeddings can surpass their initial computation cost \cite{Sun}. This chapter explores the application of Nested Embedding Models specifically utilized for Arabic natural language processing (NLP), a novel approach inspired by the hierarchical, nested structure of Matryoshka dolls.

Matryoshka Representation Learning $(MRL)$ \cite{Kusupati} is a new state-of-the-art text embedding models optimized to produce embeddings with increasingly higher output dimensions, representing input texts with more values. While traditional embeddings models \cite{Nils} produce embeddings with fixed dimensions improvement to enhance performance, it often reduces the efficiency of downstream tasks such as search or classification. Matryoshka embedding models address this issue by training embeddings to be useful even when truncated. These models can produce effective embeddings of varying dimensions.

The concept is inspired by "Matryoshka dolls", also known as "Russian nesting dolls," which are a set of wooden dolls of decreasing size placed inside one another. Similarly, Matryoshka embedding models store more critical information in the earlier dimensions and less important information in later dimensions. This characteristic allows the truncation of the original large embedding produced by the model, while still retaining sufficient information to perform well on downstream tasks. These variable-size embedding models can be highly valuable to practitioners in several ways:

\textbf{Shortlisting and Reranking}, instead of performing downstream tasks (e.g., nearest neighbor search) on the full embeddings, you can shrink the embeddings to a smaller size for efficient shortlisting. Subsequently, the remaining embeddings can be processed using their full dimensionality.

\textbf{Trade-offs}, Matryoshka models enable scaling of embedding solutions according to desired storage cost, processing speed, and performance.

The core innovation of Matryoshka Representation Learning (MRL) lies in its ability to create adaptable, nested representations through explicit optimization \cite{Kusupati}. This flexibility is crucial for large-scale classification and retrieval tasks, where computational efficiency and accuracy are paramount. 

Despite the advancements in representation learning, there has been a notable gap in the application of these sophisticated techniques to the Arabic language. Arabic, being a morphologically rich and syntactically complex language, presents unique challenges that have not been adequately addressed by existing models. This motivates the development and training of Matryoshka Embedding Models specifically for Arabic NLP downstream tasks. The main contributions of this work are summarized in three folds;

\textbf{Development of Arabic NLI Datasets}, translations of the English Stanford Natural Language Inference (SNLI) and MultiNLI datasets into Arabic using neural machine translation (NMT), providing critical resources for Arabic natural language inference (NLI) tasks.

\textbf{Training of Matryoshka Embedding Models}, training various English and Arabic embedding models, transforming them into Matryoshka versions. This process enhances their adaptability and performance across different tasks.

\textbf{Comprehensive Evaluation and Public Release}, conducting an evaluations of these trained models, offering valuable insights and making both the datasets and models publicly available on Hugging Face to facilitate broader research and application.

This chapter delves into the core principles of Matryoshka Representation Learning, highlighting its ability to create adaptable, nested representations through explicit optimization. This methodology is crucial for large-scale classification and retrieval tasks, providing significant computational benefits without compromising on accuracy. We demonstrate the practical advantages of MRL by integrating it with established NLP models, achieving notable speed-ups and maintaining high accuracy across various applications.

\section{Related Work}
In the realm of natural language processing and machine learning, representation learning has emerged as a critical area of research. The ability to create rich, multidimensional representations of data has paved the way for significant advancements in various applications, from semantic textual similarity to large-scale classification and retrieval tasks. This section reviews the key developments in representation learning, efficient classification and retrieval, and nested adaptive neural networks, highlighting the innovations that have informed the creation of Matryoshka Representation Learning (MRL). Furthermore, it positions our work within this broader context, showcasing how our contributions uniquely address the challenges and opportunities in Arabic NLP.

\textbf{Representation Learning}, the development of general-purpose representations has significantly advanced through the advent of large-scale datasets like ImageNet \cite{Deng} and JFT \cite{Sun}. These datasets enable the training of models applicable to a variety of tasks in both computer vision and natural language processing (NLP). Supervised learning typically frames representation learning as a classification problem, whereas unsupervised and self-supervised approaches employ proxy tasks such as instance discrimination and reconstruction to achieve similar goals \cite{He}. Recent breakthroughs in contrastive learning \cite{Gutmann} have further facilitated the extraction of meaningful representations from massive datasets, which are crucial for developing large-scale, cross-modal models.

Building upon these foundations, Matryoshka Representation Learning \cite{Kusupati} introduces a method to encode multiple fidelity levels within a single representation vector. This multi-fidelity approach allows for adaptive deployment, optimizing resource usage without sacrificing accuracy. The integration of $MRL$ with existing representation learning frameworks is straightforward, providing significant enhancements with minimal overhead.

\textbf{Efficient Classification and Retrieval}, efficiency in classification \cite{Harris} and retrieval \cite{Dean} is a critical concern, especially when dealing with large-scale data. Traditional methods to improve efficiency include dimensionality reduction, hashing, and feature selection. However, these techniques often reduce accuracy or increase computational complexity \cite{Waldburger}. Approximate Nearest Neighbor Search $(ANNS)$ techniques, such as Hierarchical Navigable Small World $(HNSW)$ graphs \cite{Malkov}, strike a balance between accuracy and efficiency but still face challenges related to the high dimensionality of embeddings.

MRL addresses these challenges by reducing the dimensionality of embeddings without compromising the richness of the information they encode. By nesting lower-dimensional representations within higher-dimensional ones, $MRL$ enables efficient and accurate classification and retrieval. This is particularly beneficial for adaptive systems that must operate under varying computational constraints. $MRL$’s hierarchical embeddings offer a flexible solution that scales well with data size and complexity.

\textbf{Nested and Adaptive Neural Networks}, the concept of nesting or packing neural networks of varying capacities within a larger network has been explored in the literature \cite{Cai,Li,Yu}. However, these approaches typically require separate forward passes for each nested network, which can be computationally expensive and inefficient for large-scale deployment. MRL differentiates itself by optimizing for a logarithmic number of nesting dimensions, allowing smooth interpolation between these dimensions and enabling efficient adaptive inference.

The ordered representations approach by \cite{Rippel}, which uses nested dropout in autoencoders, shares similarities with MRL but differs in its optimization strategy. MRL’s focus on coarse-to-fine granularity and minimal overhead during inference makes it particularly suited for web-scale applications. The flexibility and efficiency of MRL open new possibilities for real-time, large-scale NLP and classification tasks, demonstrating its potential as a transformative technique in the field.

The proposed work extends the principles of MRL to Arabic NLP, a domain that has not seen such sophisticated embedding techniques before. By training models specifically for Arabic using sentence transformers, we introduce the first versions of Arabic Matryoshka Embedding Models. These models not only address the unique challenges posed by the Arabic language but also provide a significant leap forward in the efficiency and adaptability of NLP systems for Arabic. Our contributions include the creation and public release of translated datasets and trained models, facilitating broader research and application in Arabic NLP. This pioneering work sets the stage for further advancements and practical implementations in this critical area.

\section{Dataset Preparation}

The preparation of datasets is a crucial step in developing robust sentence embeddings models, particularly for languages with fewer available resources such as Arabic. This section details the Arabic dataset used in this study, the translation process from English to Arabic, and the data preprocessing steps that were employed to ensure the datasets were ready for training nested embedding models.

\subsection{Arabic Dataset}

The datasets used in this study are derived from the Stanford Natural Language Inference (SNLI) \cite{Bowman} and MultiNLI \cite{Kim} datasets, which are well-known benchmarks for evaluating models on natural language inference (NLI) tasks \cite{MacCartney}. These datasets were originally designed to facilitate various NLP tasks by providing pairs of sentences along with labels that indicate their semantic relationships. To adapt these datasets for Arabic, we created multiple subsets to have more varieties in building different embeddings models.

\textbf{Pair Subset}, this subset contains pairs of sentences with the columns "anchor" and "positive." The primary purpose of this dataset is to facilitate the training of embedding models that need to learn semantic textual similarity. Table~\ref{tab:0} shows an example of the Pair subset. The dataset includes 314K training pairs, 6.81K validation pairs, and 6.83K test pairs. By using this subset, models can be trained to recognize and quantify the semantic similarity between two sentences, which is crucial for tasks such as paraphrase identification and duplicate question detection.

\begin{table}[ht]
\centering
\begin{tabular}{|c|c|c|}
\hline
\textbf{Subset} & \textbf{Anchor} & \textbf{Positive} \\
\hline
Pair & \RL{كيف أكون جيولوجياً جيداً؟} & \RL{ماذا علي أن أفعل لأكون جيولوجياً عظيماً؟} \\
     & How can I be a good geologist? & What should I do to be a great geologist? \\
\hline
\end{tabular}
\caption{Example of Arabic Pair Subset}
\label{tab:0}
\end{table}

\textbf{Triplet Subset}, this subset extends the pair subset by including a third column, "negative," to form triplets of sentences. It contains 558K training triplets, 6.58K validation triplets, and 6.61K test triplets. Table~\ref{tab:0-1} shows an example of the Triplet subset. The inclusion of a negative example allows the model to not only recognize similar pairs but also to distinguish between similar and dissimilar pairs. This is particularly useful for contrastive learning approaches, where the model learns to pull similar sentences closer and push dissimilar sentences apart in the embedding space.

\begin{table}[ht]
\centering
\begin{tabular}{|>{\centering\arraybackslash}m{0.15\textwidth}|>{\centering\arraybackslash}m{0.25\textwidth}|>{\centering\arraybackslash}m{0.25\textwidth}|>{\centering\arraybackslash}m{0.25\textwidth}|}
\hline
\textbf{Subset} & \textbf{Anchor} & \textbf{Positive} & \textbf{Negative} \\
\hline
\multirow{2}{*}{Triplet} & \RL{هناك كلب في الماء} & \RL{الكلب يسبح في بركة} & \RL{الكلب في الرمال} \\
& (There is a dog in the water) & (The dog is swimming in a pond) & (The dog is in the sand) \\
\hline
\end{tabular}
\caption{Example of Arabic Triplet Subset}
\label{tab:0-1}
\end{table}

\textbf{Pair-Class Subset}, this subset comprises three columns: "premise," "hypothesis," and "label," with 942K training examples, 19.7K validation examples, and 19.7K test examples. The label indicates the relationship between the premise and hypothesis, such as "0": "entailment", "1": "neutral", "2": "contradiction". Table~\ref{tab:0-2} shows an example of the Pair-Class subset. This subset is specifically designed for natural language inference tasks, where the goal is to determine the logical relationship between two sentences. It provides a rich resource for training models that need to understand and reason about the semantic content of sentences.

\begin{table}[ht]
\centering
\begin{tabular}{|>{\centering\arraybackslash}p{0.15\textwidth}|>{\centering\arraybackslash}p{0.35\textwidth}|>{\centering\arraybackslash}p{0.35\textwidth}|>{\centering\arraybackslash}p{0.1\textwidth}|}
\hline
\textbf{Subset} & \textbf{Premise} & \textbf{Hypothesis} & \textbf{Label} \\
\hline
\multirow{3}{*}{Pair-Class} & \RL{أطفال يبتسمون و يلوحون للكاميرا} & \RL{إنهم يبتسمون لوالديهم} & 1 \\
                            & (Children are smiling and waving at the camera) & (They are smiling at their parents) & \\
\cline{2-4}
                            & \RL{أطفال يبتسمون و يلوحون للكاميرا} & \RL{هناك أطفال حاضرون} & 0 \\
                            & (Children are smiling and waving at the camera) & (There are children present) & \\
\cline{2-4}
                            & \RL{أطفال يبتسمون و يلوحون للكاميرا} & \RL{الاطفال يتجهمون} & 2 \\
                            & (Children are smiling and waving at the camera) & (The children are frowning) & \\
\hline
\end{tabular}
\caption{Example of Arabic Pair-Class Subset}
\label{tab:0-2}
\end{table}

\textbf{Pair-Score Subset}, this subset includes columns "sentence1," "sentence2," and "score," with 942K training pairs, 19.7K validation pairs, and 19.7K test pairs. The score represents the degree of similarity between the two sentences on a continuous scale. Table~\ref{tab:0-3} shows an example of the Pair-Score subset. This subset is ideal for tasks that require a fine-grained understanding of semantic similarity, such as ranking and retrieval systems. Models trained on this subset can learn to assign similarity scores to sentence pairs, which can be used to improve the performance of search engines and recommendation systems.

\begin{table}[ht]
\centering
\begin{tabular}{|>{\centering\arraybackslash}m{0.15\textwidth}|>{\centering\arraybackslash}m{0.25\textwidth}|>{\centering\arraybackslash}m{0.25\textwidth}|>{\centering\arraybackslash}m{0.10\textwidth}|}
\hline
\textbf{Subset} & \textbf{Sentence1} & \textbf{Sentence2} & \textbf{Score} \\
\hline
\multirow{2}{*}{Pair-Score} & \RL{رجل مسن يشرب عصير البرتقال في مطعم} & \RL{رجل يشرب العصير} & \multirow{2}{*}{1} \\
& (An elderly man is drinking orange juice in a restaurant) & (A man is drinking juice) & \\
\hline
\multirow{2}{*}{Pair-Score} & \RL{عائلة أجنبية تسير على طريق ترابي بجانب الماء} & \RL{الناس يسيرون بجانب بحيرة} & \multirow{2}{*}{0.5} \\
& (A foreign family is walking on a dirt road by the water) & (People are walking by a lake) & \\
\hline
\multirow{2}{*}{Pair-Score} & \RL{عائلة أجنبية تسير على طول طريق ترابي بجانب الماء} & \RL{الناس يقودون سيارة على الطريق السريع} & \multirow{2}{*}{0} \\
& (A foreign family is walking along a dirt road by the water) & (People are driving a car on the highway) & \\
\hline
\end{tabular}
\caption{Example of Arabic Pair-Score Subset}
\label{tab:0-3}
\end{table}

\textbf{Arabic STSB Structure}, this subset is an Arabic version of the Semantic Textual Similarity Benchmark \cite{Yang}. It consists of sentence pairs drawn from diverse sources such as news headlines, video and image captions, and natural language inference data. Each pair is annotated with a similarity score normalized between 0 and 1. The dataset includes 5.75K training pairs, 1.68K validation pairs, and 1.38K test pairs. Table~\ref{tab:0-4} shows an example of the STSB Benchmark subset. This subset provides a benchmark for evaluating the performance of models on the task of semantic textual similarity in Arabic, offering a standardized way to measure and compare model performance. Table~\ref{tab:1} summarizes the subsets details with their training, validation, and test splits. 

\begin{table}[ht]
\centering
\begin{tabular}{|>{\centering\arraybackslash}m{0.12\textwidth}|>{\centering\arraybackslash}m{0.25\textwidth}|>{\centering\arraybackslash}m{0.25\textwidth}|>{\centering\arraybackslash}m{0.08\textwidth}|}
\hline
\textbf{Subset} & \textbf{Sentence1} & \textbf{Sentence2} & \textbf{Score} \\
\hline
\multirow{2}{*}{STSB Ex1} & \RL{طائرة ستقلع} & \RL{طائرة جوية ستقلع} & \multirow{2}{*}{1} \\
& (A plane is taking off) & (An airplane is taking off) & \\
\hline
\multirow{2}{*}{STSB Ex2} & \RL{رجل يعزف على ناي كبير} & \RL{رجل يعزف على الناي} & \multirow{2}{*}{0.76} \\
& (A man is playing a large flute) & (A man is playing the flute) & \\
\hline
\multirow{2}{*}{STSB Ex3} & \RL{رجل ينشر الجبن الممزق على البيتزا} & \RL{رجل ينشر الجبن الممزق على بيتزا غير مطبوخة} & \multirow{2}{*}{0.76} \\
& (A man is spreading shredded cheese on pizza) & (A man is spreading shredded cheese on an uncooked pizza) & \\
\hline
\multirow{2}{*}{STSB Ex4} & \RL{ثلاثة رجال يلعبون الشطرنج} & \RL{رجلين يلعبان الشطرنج} & \multirow{2}{*}{0.52} \\
& (Three men are playing chess) & (Two men are playing chess) & \\
\hline
\end{tabular}
\caption{Examples of Arabic STSB Benchmark Subset}
\label{tab:0-4}
\end{table}

\begin{table}[ht]
\centering
\resizebox{\textwidth}{!}{%
\begin{tabular}{|c|c|c|c|c|}
\hline
Subset            & Columns                           & Training Examples & Validation Examples & Test Examples \\ \hline
Pair Subset       & "anchor", "positive"              & 314K              & 6.81K               & 6.83K         \\ \hline
Triplet Subset    & "anchor", "positive", "negative"  & 558K              & 6.58K               & 6.61K         \\ \hline
Pair-Class Subset & "premise", "hypothesis", "label"  & 942K              & 19.7K               & 19.7K         \\ \hline
Pair-Score Subset & "sentence1", "sentence2", "score" & 942K              & 19.7K               & 19.7K         \\ \hline
Arabic STSB Structure & "sentence1", "sentence2", "similarity score" & 5.75K & 1.68K & 1.38K \\ \hline
\end{tabular}%
}
\caption{Arabic NLI Subsets Details}
\label{tab:1}
\end{table}

These subsets cover a broad range of tasks from semantic textual similarity to classification, making them versatile for training and evaluating embedding models.

\subsection{Data Preprocessing}
Data preprocessing is a critical step to prepare the raw translated text for model training. The following preprocessing steps were applied to both the original Arabic and the translated datasets:

\textbf{Tokenization}, the text was tokenized into individual words or sub-words using SentencePiece \cite{Kudo}, which helps in handling the morphological richness of Arabic more effectively. This step breaks down the text into smaller units, making it easier for the model to learn and generalize from the data.

\textbf{Normalization}, text normalization was performed to standardize various forms of Arabic script. This included the removal of diacritics, normalization of character forms, and handling of punctuation. Normalization ensures that different forms of the same word are treated uniformly by the model.

\textbf{Data Structuring}, the datasets were structured into the required format with specific columns for each task. For instance, pairs of sentences were organized into "anchor" and "positive" columns for similarity tasks, while triplets included an additional "negative" column. This structuring helps in efficiently loading and processing the data during training.

\textbf{Validation and Test Splits}, the datasets were split into training, validation, and test sets to ensure proper evaluation of the models. The splits were carefully maintained to ensure that the distributions of data remained consistent across these sets. This step is crucial for evaluating the performance of the models and preventing overfitting.

\textbf{Saving Processed Data}, the processed datasets were saved in CSV format to facilitate easy loading and use in subsequent training phases. The data was encoded in $UTF-8$ to preserve the integrity of Arabic characters. Storing the data in a structured format ensures that it can be easily shared and reused by other researchers.

By obtaining these preprocessing steps, we prepared the datasets for the machine translation process to translate these subsets into Arabic language. 

\subsection{Translation Process}
The translation of the datasets from English to Arabic was performed using Neural Machine Translation (NMT) \cite{Klein}, a sophisticated technique that leverages neural networks to achieve high-quality translations. This process involved several meticulous steps to ensure the quality and accuracy of the translations, which are crucial for effective downstream NLP tasks.

\textbf{Dataset Loading}, the first step involved loading the English datasets using the Hugging Face datasets library\footnote{https://huggingface.co/docs/datasets/en/index}. This library offers a seamless way to access and manipulate a variety of NLP datasets. By leveraging this library, we ensured that the data was in a structured format, making it easier to process and translate. The datasets included various subsets such as the SNLI and MultiNLI datasets, which are benchmark datasets for natural language inference.

\textbf{Translation Model Configuration}, for the translation task, we used the CTranslate2 model\footnote{https://github.com/OpenNMT/CTranslate2}, a powerful NMT model designed for efficient translation tasks. To handle sub-word tokenization, we employed the SentencePiece model. SentencePiece is particularly useful for languages like Arabic, which have rich morphological structures. The model paths for CTranslate2 and SentencePiece were set to ensure that the models could correctly process the source and target languages.

\textbf{Language Specification}, it was essential to correctly specify the source and target languages to ensure accurate translations. We used language codes $eng\_Latn$ for English and $arb\_Arab$ for Arabic. This specification helped the NMT model to understand the linguistic characteristics of both languages, thereby improving the quality of the translations.

\textbf{Batch Translation}, given the large size of the datasets, the translation process was performed in batches. This approach helped manage computational resources effectively and ensured that the translation process was scalable. Each batch of sentences was first tokenized into sub-words using SentencePiece. This tokenization step is crucial because it breaks down the text into manageable units, which the NMT model can then process more effectively.

\textbf{Handling Special Tokens}, special tokens, such as language tags and end-of-sequence markers, were handled carefully to avoid any artifacts in the translated text. For example, the source sentences were prefixed with the source language tag, and the target sentences were postfixed with the target language tag. After translation, any unnecessary tokens were removed to ensure that the final output was clean and ready for downstream tasks.

These steps of translation are conducted and adapted for each on the data subsets with their columns ensuring translating almost the same number of samples as in original. Moreover, the translation process steps are further controlled by the following main steps including;

\textbf{SentencePiece Tokenization}, each sentence was tokenized into sub-words using SentencePiece. This step ensured that the NMT model could handle the text more efficiently, especially for languages with complex morphology like Arabic.

\textbf{Translation with CTranslate2}, the tokenized sentences were then fed into the CTranslate2 model for translation. The model produced translated sentences in the form of sub-word tokens.

\textbf{Detokenization}, the sub-word tokens were then detokenized back into full sentences using SentencePiece. This step is crucial to reconstruct the sentences in the target language accurately.

Additional steps have been taken into consideration as well focusing on maintaining data consistency and quality assurance where the translated dataset was structured to maintain consistency with the original dataset format. This consistency is vital for ensuring that the translated datasets can be easily integrated into downstream tasks without requiring significant modifications. The structure included specific columns for different types of tasks, such as sentence pairs for similarity tasks and triplets for contrastive learning tasks. Moreover, to ensure the quality of the translations, we performed several checks. A random sample of the translated sentences was manually reviewed to verify the accuracy and fluency of the translations. 

The meticulous translation process described above resulted in high-quality Arabic versions of the SNLI and MultiNLI datasets. These translated datasets are structured to facilitate various NLP tasks and are made publicly available on Hugging Face\footnote{https://huggingface.co/collections/Omartificial-Intelligence-Space/arabic-nli-and-semantic-similarity-datasets-6671ba0a5e4cd3f5caca50c3} to enable broader research and development in Arabic NLP. This translation process not only ensures high-quality data but also sets a precedent for translating other important NLP datasets into Arabic or other low-resource languages.

The combined efforts in dataset preparation, translation, and preprocessing resulted in a robust set of Arabic NLP datasets that are versatile and ready for training state-of-the-art embedding models. The public release of these datasets on Hugging Face will enable broader research and development in Arabic NLP, fostering advancements in this critical area.

\section{Methodology}

The methodology section outlines the steps and processes involved in the development and evaluation of the Arabic Matryoshka Embedding Models. This includes the selection of appropriate models, training procedures, and evaluation techniques used to assess their performance on Arabic NLP tasks.

\subsection{Model Selection}

In this study, we aimed to train and evaluate several Matryoshka embedding models using the Arabic $NLI$ triplet dataset. The selection of models was guided by their proven effectiveness in various NLP tasks, as well as their ability to handle Arabic text to varying extents. The models chosen for this investigation include both monolingual and multilingual sentence transformers, as well as models specifically designed for the Arabic language.

\textbf{English Sentence Transformer Model}, we have utilized $mpnet-base-all-nli-triplet$\footnote{https://huggingface.co/tomaarsen/mpnet-base-all-nli-triplet} model that is a fine-tuned version from $microsoft/mpnet-base$\footnote{https://huggingface.co/microsoft/mpnet-base}, which maps sentences and paragraphs to a 768-dimensional dense vector space. It is designed for tasks such as semantic textual similarity, semantic search, paraphrase mining, text classification, clustering, and more. Despite being primarily trained on English data, it has been exposed to a few Arabic tokens, making it a suitable candidate for assessing cross-lingual transfer capabilities in our investigation.

\textbf{Multilingual Sentence Transformer Models}, we have utilized $paraphrase-multilingual-mpnet-base-v2$\footnote{https://huggingface.co/sentence-transformers/paraphrase-multilingual-mpnet-base-v2} model that maps sentences and paragraphs to a 768-dimensional dense vector space and is suitable for tasks like clustering and semantic search. It supports multiple languages, making it a strong candidate for multilingual NLP applications. Additionally, we have used $LaBSE$ (Language-Agnostic BERT Sentence Embedding)\footnote{https://huggingface.co/sentence-transformers/LaBSE}, model that maps 109 languages including Arabic into a shared vector space, facilitating cross-lingual tasks. It is a robust choice for evaluating multilingual capabilities in embedding models.

\textbf{Arabic Sentence Transformer Models}, to expand our investigation, we have used Arabic based sentence transformers including $AraBERT$\footnote{https://huggingface.co/aubmindlab/bert-base-arabertv02}, an arabic pretrained language model based on Google's BERT architecture. It uses the BERT-Base configuration and is tailored for Arabic NLP tasks. AraBERT has demonstrated high performance in various Arabic language benchmarks. Moreover, $MARBERT$\footnote{https://huggingface.co/UBC-NLP/MARBERT} model which is designed to handle both Modern Standard Arabic (MSA) and Dialectal Arabic (DA). It is based on the BERT architecture and trained on a large corpus of Arabic tweets, making it particularly effective for tasks involving social media text.

The choice of these models allows for a comprehensive evaluation across different dimensions of Arabic NLP. The inclusion of an English model exposed to some Arabic tokens provides insights into the transferability of learned representations across languages. The multilingual models offer a perspective on how well models trained on diverse language data perform on Arabic specific tasks. Finally, the Arabic-specific models, $AraBERT$ and $MARBERT$, serve as benchmarks for performance in native Arabic NLP contexts.

By training  these models on the Arabic $NLI$ triplet dataset, we aim to create Matryoshka embedding versions that are not only versatile and efficient but also tailored to the unique characteristics and challenges of the Arabic language. This thorough approach ensures that our models are well-equipped to handle a variety of NLP tasks, including semantic textual similarity, semantic search, paraphrase mining, text classification, and clustering, thereby advancing the state of Arabic NLP.

\begin{figure}[ht]
\centering
\includegraphics[width=\textwidth]{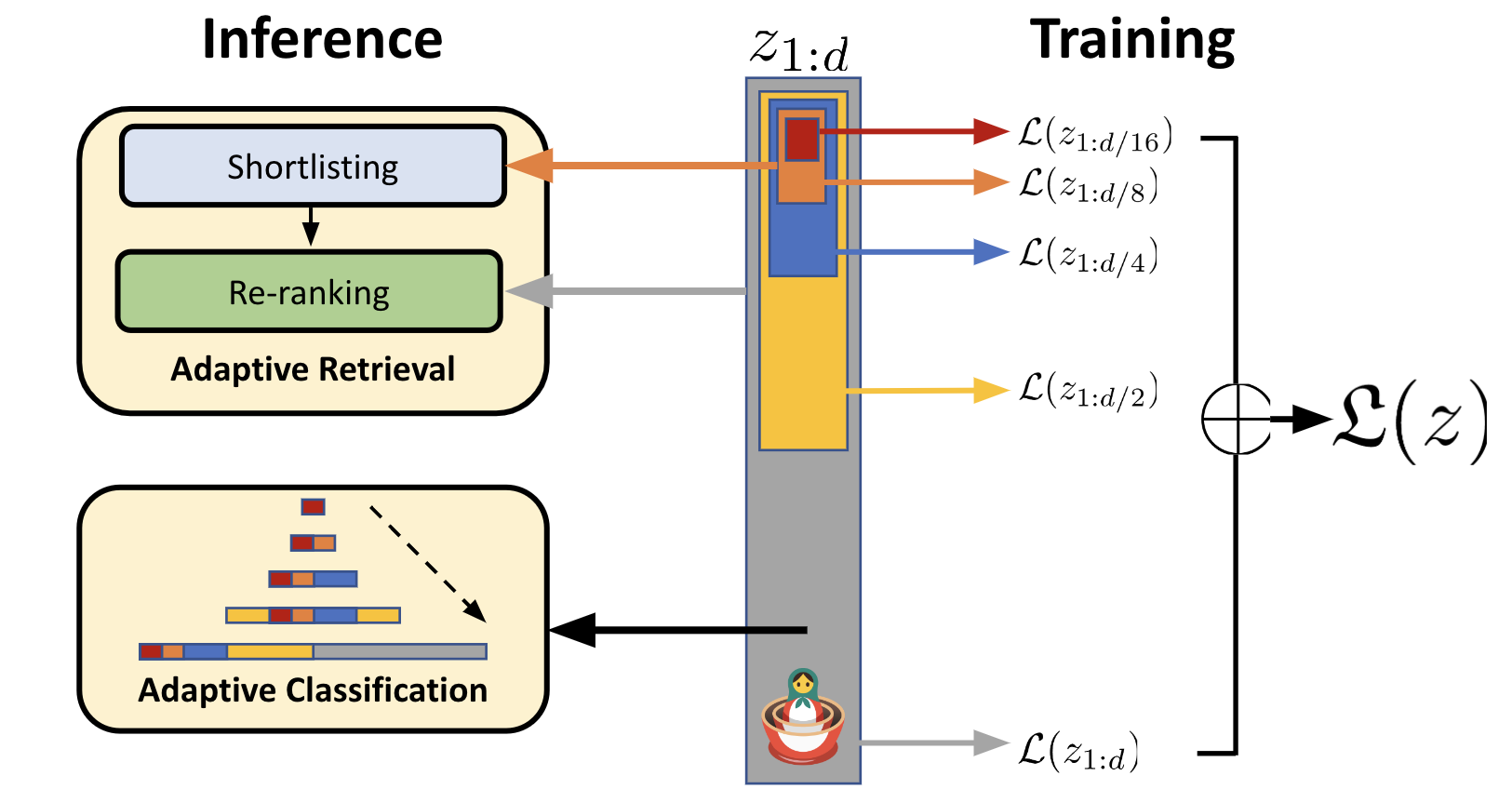}
\caption{Matryoshka Representation Learning Process \cite{Kusupati}}
\label{fig:0}
\end{figure}

\subsection{Matryoshka Embedding Models}

Matryoshka Embedding Models \cite{Kusupati} represent an innovative approach to creating adaptable and efficient embeddings for natural language processing tasks. These models aim to capture multi-granularity in embeddings, allowing different levels of the embedding vector to independently serve as meaningful representations. This adaptability is crucial for efficiently managing computational resources, especially in large-scale and resource-constrained environments like Arabic NLP.

Matryoshka Representation Learning $(MRL)$ process is shown in Figure~\ref{fig:0} which involves a d-dimensional representation vector $z \in R^{d}$ for a given datapoint $x$ in the input domain $\chi$. This representation vector is obtained using a deep neural network $F\left(.;\theta_{F} \right)$, parameterized by learnable weights $\theta_{F}$. The objective of MRL is to ensure that each of the first $m$ dimensions of the embedding vector, $z_{1:m} \in R^{m}$, where $m\in M$, can independently serve as a transferable and general-purpose representation of the datapoint $x$.

The multi-granularity of these embeddings is captured through the set of chosen dimensions $M$, which are determined by consistent halving until the representation size reaches a minimal informative state. This nested, coarse-to-fine granularity ensures that the representations remain useful even when truncated to smaller dimensions.

Given a labeled dataset $D = \left\{ \left( x_{1} , y_{1} \right) , ...,\left( x_{N}, y_{N} \right) \right\}$, where $x_{i} \in \chi$ is an input point and $y_{i} \in \left[ L \right]$ represents the label of $x_{i}$ MRL optimizes the multi-class classification loss for each nested dimension $m\in M$ using standard empirical risk minimization. This is achieved by employing a separate linear classifier, parameterized by $\textbf{W}^{\left( m \right)}\in R^{L \times m}$ for each dimension. The losses obtained from these classifiers are then aggregated, taking into account their relative importance $c_{m} \ge 0$ for $m\in M$.  The optimization objective can be expressed as:

\begin{equation} \label{eq1}
\mathcal{L}_{\text{MRL}} = \sum_{m \in M} c_m \mathcal{L}_{\text{CE}}(\mathbf{W}^{(m)} \mathbf{z}_{1:m}, y)
\end{equation}

where:
\begin{itemize}
  \item \( \mathcal{L}_{\text{MRL}} \) is the Matryoshka Representation Learning loss.
  \item \( c_m \) represents the relative importance of each dimension \( m \).
  \item \( \mathcal{L}_{\text{CE}} \) is the multi-class softmax cross-entropy loss function.
  \item \( \mathbf{W}^{(m)} \in \mathbb{R}^{L \times m} \) are the weights of the linear classifier for dimension \( m \).
  \item \( \mathbf{z}_{1:m} \in \mathbb{R}^m \) is the truncated embedding vector up to dimension \( m \).
  \item \( y \) is the true label corresponding to the input \( x \).
\end{itemize}

This formulation ensures that each subset of the embedding dimensions can independently perform well on the classification task, maintaining the flexibility and robustness of the learned representations.

To enhance efficiency, weight-tying is employed across all the linear classifiers, i.e., defining $\textbf{W}^{\left( m \right)} = \textbf{W}_{1:m}$ for a set of common weights $\textbf{W}$. This reduces the memory cost associated with the linear classifiers, which is particularly crucial in cases of extremely large output spaces. This variant is known as Efficient Matryoshka Representation Learning $(MRL-E)$.

In practice, MRL involves the following steps:

\textbf{Representation Vector Generation}, a deep neural network generates a high-dimensional representation vector for each input datapoint.

\textbf{Dimension Nesting}, the high-dimensional vector is divided into nested subsets of dimensions, ensuring that the first few dimensions capture the most crucial information.

\textbf{Separate Classifier Training}, each subset of dimensions is used to train a separate linear classifier, optimizing the classification loss for each dimension.

\textbf{Loss Aggregation}, the classification losses from all subsets are aggregated based on their relative importance to form the final loss function.

By leveraging these methods, Matryoshka Embedding Models achieve flexibility and efficiency, making them suitable for adaptive deployment in various NLP tasks. Figure~\ref{0-2} shows the truncation step of the Matryoshka embedding where the ability to truncate embeddings is performed without significant loss of information allows for scalable and resource-efficient applications, particularly beneficial in the context of Arabic NLP where computational resources may be limited.

\begin{figure}[ht]
\centering
\includegraphics[width=\textwidth]{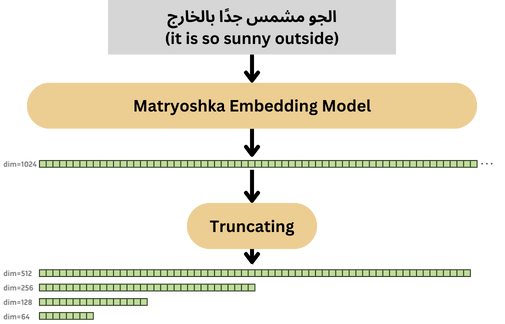}
\caption{Truncation Step in Matryoshka Representation Learning}
\label{fig:0-2}
\end{figure}

\subsection{Nested Embedding Training Process}

The process of training nested embedding models on the translated Arabic dataset was methodically structured to ensure optimal performance. The training setup began with the initialization of several Sentence Transformer models. The primary model used was $mpnet-base-all-nli-triplet$, which maps sentences and paragraphs to a 768-dimensional dense vector space. Additionally, two multilingual sentence transformer models, $paraphrase-multilingual-mpnet-base-v2$ and $LaBSE$, and two Arabic-specific models, $AraBERT$ and $MARBERT$, were trained. These models were selected to leverage their existing capabilities while adapting them to handle the specific characteristics of the Arabic language.

The dataset used for training, the $arabic-nli-triplet$ subset, was loaded using the Hugging Face datasets library. This dataset comprises triplets of sentences, each consisting of an anchor, a positive, and a negative example with size of 558k samples. The models were configured to handle these inputs, with settings such as using cosine similarity as the similarity function and defining the maximum sequence length as 512 tokens.

Hyper-parameter tuning was a critical aspect of the training process. The batch size was set to 128 to balance the computational load and model performance, and each model was trained for one epoch, leveraging the computational power of an A100 GPU. The Matryoshka embedding approach was implemented by specifying output dimensions of [768, 512, 256, 128, 64] , allowing the models to produce embeddings at various levels of granularity. This hierarchical structure enables flexible and efficient processing of downstream tasks.

Optimization during training involved the use of specific loss functions. The primary loss function, $MultipleNegativesRankingLoss$ \cite{Henderson}, was augmented with $MatryoshkaLoss$ to train embeddings at multiple dimensions simultaneously \cite{Kusupati}. This combined approach ensured that the embeddings were effective across different levels of granularity, making them robust for various applications.

The training process was managed by data loading and preprocessing involving shuffling the dataset and selecting a subset of examples for training to manage computational resources effectively. Each triplet was processed to ensure consistency and quality. The $MultipleNegativesRankingLoss$ handled the ranking of the embeddings, while the $MatryoshkaLoss$ maintained their hierarchical structure. An evaluator was set up using the $STS$ Benchmark validation dataset to monitor performance during training. This involved calculating cosine similarity between sentence pairs and tracking performance across different dimensions of the embeddings.

The training execution was conducted using $SentenceTransformerTrainer$ from the Sentence Transformers library\footnote{https://sbert.net/index.html}. The training arguments included specifications for the number of epochs, batch size, learning rate scheduling, and evaluation strategies. After training, the models were evaluated on the $STS$ Benchmark test dataset to assess their performance across different dimensions. The results were tracked and recorded, highlighting the effectiveness of the Matryoshka embedding approach. Finally, the trained models were saved locally and uploaded to the Hugging Face Hub for public access\footnote{https://huggingface.co/collections/Omartificial-Intelligence-Space/arabic-matryoshka-embedding-models-666f764d3b570f44d7f77d4e}.

This meticulous training ensured that the process Matryoshka embedding models were robust, efficient, and ready for deployment in various NLP tasks, particularly in the context of Arabic language processing.

\section{Results \& Discussion}

In this section, we provide a detailed framework of the evaluation process and the metrics used to assess the performance of the trained Arabic Matryoshka embedding models on Arabic datasets. Our aim is to thoroughly examine the models' efficacy in various dimensions and discuss the outcomes in the context of Arabic natural language processing tasks.

To evaluate the performance of our models, we utilized the Arabic Semantic Textual Similarity Benchmark $(STSB)$\footnote{https://huggingface.co/datasets/Omartificial-Intelligence-Space/Arabic-stsb}. This dataset is a comprehensive collection of sentence pairs drawn from diverse sources, including news headlines, video and image captions, and natural language inference data. Each sentence pair in the dataset is human-annotated with a similarity score ranging from 1 to 5, which we normalized to a scale between 0 and 1 for this evaluation. This normalization ensures consistency and facilitates a more straightforward comparison across different models and dimensions.

The evaluation of our Arabic Matryoshka embedding models was conducted using the \\$EmbeddingSimilarityEvaluator$, a robust evaluation tool provided by the Sentence Transformers library. This evaluator is specifically designed to measure the similarity of embeddings by calculating the Spearman and Pearson rank correlation coefficients against the gold standard labels. These correlations provide a quantitative measure of how well the embeddings capture the semantic similarity between sentence pairs.

The $EmbeddingSimilarityEvaluator$ performs evaluation based on several metrics:

\textbf{Cosine Similarity}, measures the cosine of the angle between two vectors, providing a value between -1 and 1.

\textbf{Manhattan Distance}, calculates the absolute differences between corresponding elements of the vectors.

\textbf{Euclidean Distance}, computes the straight line distance between two points in the vector space.

\textbf{Dot Product Similarity}, computes the dot product of two vectors, which can indicate similarity in direction and magnitude.

These metrics were computed for each dimension of the embeddings \textsl{(768, 512, 256, 128, 64)}, 
resulting in a comprehensive evaluation of the models' performance at various levels of granularity. 
For each dimension, we used the following metrics:

\textbf{Pearson Correlation (Cosine, Manhattan, Euclidean, Dot, Max)}, measures the linear correlation between the predicted and actual similarity scores.

\textbf{Spearman Correlation (Cosine, Manhattan, Euclidean, Dot, Max)}, Assesses the monotonic relationship between the predicted and actual similarity scores.

The evaluation process commenced with loading the pretrained Sentence Transformer models and the $STSB$ dataset using the Hugging Face datasets library. Each model was evaluated on the sentence pairs in the dataset, with the $EmbeddingSimilarityEvaluator$ applied to calculate the similarity scores. This process involved initializing the evaluator with the sentences and their corresponding human-annotated scores, specifying the primary similarity function (cosine similarity), and setting other relevant parameters such as batch size and truncation dimension.

The evaluator outputs a set of correlation scores for each metric and dimension, providing a detailed view of how well the embeddings align with human judgment of sentence similarity. This methodology ensures a rigorous and standardized evaluation, enabling us to compare the performance of different models and configurations objectively.

The results from this evaluation process will be discussed in the subsequent subsections, where we will delve into the performance metrics, compare the models, and interpret the findings in the context of their application to Arabic NLP tasks. This detailed analysis will highlight the strengths and potential areas for improvement in our Arabic Matryoshka embedding models, contributing to the broader understanding and development of efficient and effective NLP solutions for the Arabic language.

\subsection{Comprehensive Performance Analysis For Each Trained Arabic Matryoshka Embedding Model}

A deep investigation have been done to evaluate the performance of different trained Arabic Matryoshka embedding models on different dimensions. The experiments consists of evaluating Arabic embeddings models and multilingual embeddings models along with An English model subjected to a small number of Arabic tokens is included to shed light on the cross-linguistic transferability of learnt representations. Results of the first multilingual model named  $Paraphrase-Multilingual-MPNet-Base-V2$ which is trained to create Arabic Matryoshka embeddings across different dimensions are summarized in Table~\ref{tab:2}.

\begin{table}[]
\centering
\resizebox{\columnwidth}{!}{%
\begin{tabular}{|c|c|c|c|c|c|c|c|c|c|c|}
\hline
Dimension & Pearson Cosine & Spearman Cosine & Pearson Manhattan & Spearman Manhattan & Pearson Euclidean & Spearman Euclidean & Pearson Dot & Spearman Dot & Pearson Max & Spearman Max \\ \hline
768       & 0.8539         & 0.8616          & 0.8497            & 0.8513             & 0.8516            & 0.8541             & 0.7281      & 0.7230       & 0.8539      & 0.8616       \\ \hline
512       & 0.8542         & 0.8609          & 0.8487            & 0.8512             & 0.8505            & 0.8539             & 0.7076      & 0.7029       & 0.8542      & 0.8609       \\ \hline
256       & 0.8486         & 0.8579          & 0.8405            & 0.8456             & 0.8415            & 0.8472             & 0.6669      & 0.6651       & 0.8486      & 0.8579       \\ \hline
128       & 0.8390         & 0.8499          & 0.8287            & 0.8353             & 0.8298            & 0.8372             & 0.5856      & 0.5835       & 0.8390      & 0.8499       \\ \hline
64        & 0.8291         & 0.8429          & 0.8101            & 0.8221             & 0.8129            & 0.8255             & 0.5067      & 0.5110       & 0.8291      & 0.8429       \\ \hline
\end{tabular}%
}
\caption{Performance of Trained Paraphrase Multilingual MPNet Base V2 Matryoshka Model}
\label{tab:2}
\end{table}

As shown in Table~\ref{tab:2}, the model shows robust performance across all dimensions, with a slight decrease in correlation as the dimensionality reduces. The highest Pearson and Spearman correlations are observed at dimensions 768 and 512, indicating strong semantic similarity capture. Moreover, Dot product similarity metrics decrease significantly with lower dimensions, highlighting its sensitivity to dimensionality reduction.

Secondly, results of the second multilingual model 'LaBSE' are detailed in Table~\ref{tab:3}.

\begin{table}[]
\centering
\resizebox{\columnwidth}{!}{%
\begin{tabular}{|c|c|c|c|c|c|c|c|c|c|c|}
\hline
Dimension & Pearson Cosine & Spearman Cosine & Pearson Manhattan & Spearman Manhattan & Pearson Euclidean & Spearman Euclidean & Pearson Dot & Spearman Dot & Pearson Max & Spearman Max \\ \hline
768       & 0.7269         & 0.7225          & 0.7259            & 0.721              & 0.726             & 0.7225             & 0.7269      & 0.7225       & 0.7269      & 0.7225       \\ \hline
512       & 0.7268         & 0.7224          & 0.7241            & 0.7195             & 0.7248            & 0.7213             & 0.7253      & 0.7205       & 0.7268      & 0.7224       \\ \hline
256       & 0.7283         & 0.7264          & 0.7228            & 0.7181             & 0.7251            & 0.7215             & 0.7243      & 0.7221       & 0.7283      & 0.7264       \\ \hline
128       & 0.7102         & 0.7104          & 0.7135            & 0.7089             & 0.7172            & 0.713              & 0.6778      & 0.6746       & 0.7172      & 0.713        \\ \hline
64        & 0.6931         & 0.6982          & 0.6971            & 0.6942             & 0.7013            & 0.6987             & 0.6377      & 0.6345       & 0.7013      & 0.6987       \\ \hline
\end{tabular}%
}
\caption{Performance of Trained LaBSE Matryoshka Model}
\label{tab:3}
\end{table}

As shown in Table~\ref{tab:3}, LaBSE exhibits high performance across all dimensions, showing minimal degradation in correlation values as dimensions decrease. The model's robustness is evident in maintaining high Pearson and Spearman correlations across various metrics.

Moving on, Tables~\ref{tab:4} and ~\ref{tab:5} shows the performance evaluation metrics for the arabic embeddings models Arabert and MArbert respecitvely.

\begin{table}[]
\centering
\resizebox{\columnwidth}{!}{%
\begin{tabular}{|c|c|c|c|c|c|c|c|c|c|c|}
\hline
Dimension & Pearson Cosine & Spearman Cosine & Pearson Manhattan & Spearman Manhattan & Pearson Euclidean & Spearman Euclidean & Pearson Dot & Spearman Dot & Pearson Max & Spearman Max \\ \hline
768       & 0.595          & 0.616           & 0.6296            & 0.627              & 0.6327            & 0.6317             & 0.4282      & 0.4295       & 0.6327      & 0.6317       \\ \hline
512       & 0.5846         & 0.6064          & 0.6288            & 0.6264             & 0.6313            & 0.6302             & 0.3789      & 0.3768       & 0.6313      & 0.6302       \\ \hline
256       & 0.5779         & 0.596           & 0.6243            & 0.6217             & 0.6238            & 0.6215             & 0.3597      & 0.353        & 0.6243      & 0.6215       \\ \hline
128       & 0.5831         & 0.6022          & 0.6152            & 0.6122             & 0.6162            & 0.6153             & 0.4044      & 0.4015       & 0.6162      & 0.6153       \\ \hline
64        & 0.5725         & 0.5914          & 0.6024            & 0.5967             & 0.6069            & 0.6041             & 0.3632      & 0.3585       & 0.6069      & 0.6041       \\ \hline
\end{tabular}%
}
\caption{Performance of Trained AraBERT Matryoshka Model}
\label{tab:4}
\end{table}

\begin{table}[]
\centering
\resizebox{\columnwidth}{!}{%
\begin{tabular}{|c|c|c|c|c|c|c|c|c|c|c|}
\hline
Dimension & Pearson Cosine & Spearman Cosine & Pearson Manhattan & Spearman Manhattan & Pearson Euclidean & Spearman Euclidean & Pearson Dot & Spearman Dot & Pearson Max & Spearman Max \\ \hline
768       & 0.6112         & 0.6117          & 0.6444            & 0.6358             & 0.6444            & 0.6346             & 0.4724      & 0.4484       & 0.6444      & 0.6358       \\ \hline
512       & 0.6665         & 0.6648          & 0.643             & 0.6335             & 0.6466            & 0.6373             & 0.537       & 0.5242       & 0.6665      & 0.6648       \\ \hline
256       & 0.6601         & 0.6593          & 0.6362            & 0.6251             & 0.6408            & 0.63               & 0.5251      & 0.5155       & 0.6601      & 0.6593       \\ \hline
128       & 0.6549         & 0.6523          & 0.6343            & 0.6227             & 0.6397            & 0.6281             & 0.4724      & 0.4634       & 0.6549      & 0.6523       \\ \hline
64        & 0.6367         & 0.637           & 0.6264            & 0.6119             & 0.6328            & 0.618              & 0.4117      & 0.4044       & 0.6367      & 0.637        \\ \hline
\end{tabular}%
}
\caption{Performance of Trained Marbert Matryoshka Model}
\label{tab:5}
\end{table}

As shown in Table~\ref{tab:4}, Trained Arabert Matryoshka Model shows moderate performance, with correlations improving slightly as dimensions decrease from 768 to 64. While the results are lower than those for multilingual models, they are consistent across various metrics. Moreover, as shown in Table~\ref{tab:5} for Marbert model shows strong performance at higher dimensions, with consistent Pearson and Spearman correlations. The model's performance in dot product similarity is lower compared to other metrics, suggesting an area for potential improvement.

Finally, the results of English model $MPNet-Base-All-NLI-Triplet$, which has seen a few Arabic tokens, are shown in Table~\ref{tab:6}.

\begin{table}[]
\centering
\resizebox{\columnwidth}{!}{%
\begin{tabular}{|c|c|c|c|c|c|c|c|c|c|c|}
\hline
Dimension & Pearson Cosine & Spearman Cosine & Pearson Manhattan & Spearman Manhattan & Pearson Euclidean & Spearman Euclidean & Pearson Dot & Spearman Dot & Pearson Max & Spearman Max \\ \hline
768       & 0.6699         & 0.6757          & 0.6943            & 0.684              & 0.6973            & 0.6873             & 0.5534      & 0.5422       & 0.6973      & 0.6873       \\ \hline
512       & 0.6628         & 0.6703          & 0.6917            & 0.6816             & 0.6949            & 0.6853             & 0.5229      & 0.5114       & 0.6949      & 0.6853       \\ \hline
256       & 0.6368         & 0.6513          & 0.6832            & 0.6746             & 0.6844            & 0.676              & 0.4266      & 0.4179       & 0.6844      & 0.676        \\ \hline
128       & 0.6148         & 0.6355          & 0.6731            & 0.6653             & 0.6764            & 0.6691             & 0.3513      & 0.3445       & 0.6764      & 0.6691       \\ \hline
64        & 0.5789         & 0.6081          & 0.6579            & 0.6519             & 0.663             & 0.6571             & 0.2403      & 0.2331       & 0.663       & 0.6571       \\ \hline
\end{tabular}%
}
\caption{Performance of Trained MPNet-Base-All-NLI-Triplet Matryoshka Model}
\label{tab:6}
\end{table}

As shown in Table~\ref{tab:6}, the performance of the MPNet-Base-All-NLI-Triplet Matryoshka Model is noticeably lower compared to the multilingual model, especially in lower dimensions. Despite this, the model shows reasonable consistency across Pearson and Spearman correlations. The ability to handle Arabic tokens, though limited, provides some insights into the adaptability of English-trained models.

The evaluation results indicate that multilingual models like $Paraphrase-Multilingual-MPNet-Base-V2$ and $LaBSE$ outperform specifically Arabic-trained models in capturing semantic similarity in Arabic text. The trained $Paraphrase-Multilingual-MPNet-Base-V2$ model demonstrated the highest performance, particularly in higher dimensions. In contrast, models specifically designed for Arabic, such as $BERT-Base-AraBERTV02 $and $MARBERT$, show moderate performance but provide valuable insights into the nuances of Arabic language processing. The decrease in performance at lower dimensions for all models highlights the challenge of dimensionality reduction in maintaining semantic integrity.

Overall, these results underscore the importance of multilingual capabilities in embedding models for diverse language tasks and point to areas where Arabic-specific models can be further improved to match their multilingual counterparts.

\subsection{Comparative Analysis of Different Arabic Trained Matryoshka Models Performance Across Metrics and Dimensions}

\begin{figure}[ht]
    \centering
    \includegraphics[width=\textwidth]{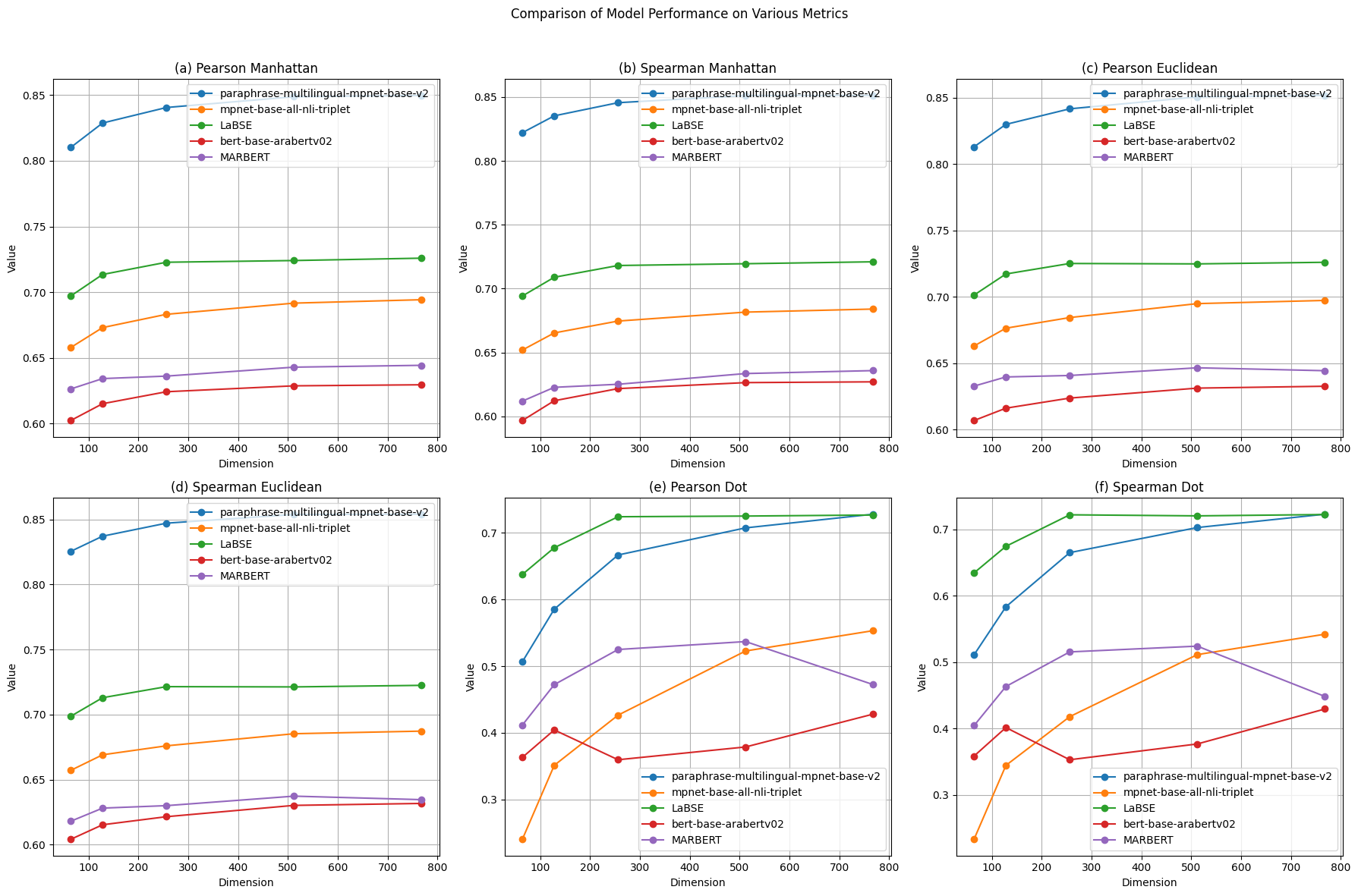}
    \caption{Comparative Analysis of Model Performance Across Different Metrics and Dimensions.}
    \label{fig:1}
\end{figure}

This subsection provides a detailed visual comparison of the performance of various Arabic Matryoshka embedding models across different dimensions. By examining the comparative plots for multiple evaluation metrics, including Pearson and Spearman correlations, we gain insights into how each model performs at different levels of granularity. Plots shown in Figure~\ref{fig:1} serve as a visual representation of the quantitative data, offering a clearer perspective on the relative performance of each model.

\textbf{Comparison of Model Performance on Pearson Manhattan Similarity} is shown in Figure~\ref{fig:1} (a) where, the $paraphrase-multilingual-mpnet-base-v2$ model consistently outperformed other models across all dimensions, exhibiting the highest Pearson Manhattan similarity values. $LaBSE$ demonstrated stable performance with slight decreases at lower dimensions. The $mpnet-base-all-nli-triplet$ model showed moderate performance, surpassing the Arabic-specific models, $bert-base-arabertv02$, and MARBERT. Among these, $MARBERT$ and $bert-base-arabertv02$ had the lowest values, indicating less effectiveness in capturing semantic similarity using Pearson Manhattan similarity.

\textbf{Comparison of Model Performance on Spearman Manhattan Similarity}, is shown in Figure~\ref{fig:1} (b) where, the $paraphrase-multilingual-mpnet-base-v2$ model again showed the highest Spearman Manhattan similarity values, reinforcing its strong performance. $LaBSE$ followed closely, maintaining stable performance across dimensions. The $mpnet-base-all-nli-triplet$ model's performance was moderate, similar to its Pearson Manhattan metric results. $MARBERT$ and $bert-base-arabertv02$ exhibited lower values, suggesting less consistency in semantic similarity capture.

\textbf{Comparison of Model Performance on Pearson Euclidean Similarity} is shown in Figure~\ref{fig:1} (c) where, the Pearson Euclidean similarity is compared, the $paraphrase-multilingual-mpnet-base-v2$ model maintained high values across dimensions. $LaBSE$ also performed consistently well. The $mpnet-base-all-nli-triplet$ model showed moderate performance, surpassing the Arabic-specific models. $MARBERT$ and $bert-base-arabertv02$ showed lower values, with $MARBERT$ slightly outperforming $bert-base-arabertv02$.

\textbf{Comparison of Model Performance on Spearman Euclidean Similarity}, is shown in Figure~\ref{fig:1} (d) for Spearman Euclidean similarity where, the $paraphrase-multilingual-mpnet-base-v2$ model led with the highest values. $LaBSE$ followed closely, showing stable values across dimensions. The $mpnet-base-all-nli-triplet$ model demonstrated moderate performance. The lowest values were observed in $MARBERT$ and $bert-base-arabertv02$, with $MARBERT$ slightly outperforming $bert-base-arabertv02$.

\textbf{Comparison of Model Performance on Pearson Dot Similarity}, is shown in Figure~\ref{fig:1} (e) where, The performance of the $paraphrase-multilingual-mpnet-base-v2$ model decreased with lower dimensions, showing the highest Pearson Dot similarity value at 768 dimensions. $LaBSE$ exhibited a similar trend, with noticeable drops at lower dimensions. The $mpnet-base-all-nli-triplet$ model showed moderate performance, with a significant decline at lower dimensions. $MARBERT$ and $bert-base-arabertv02$ had the lowest values, with $MARBERT$ slightly outperforming $bert-base-arabertv02$.

\textbf{Comparison of Model Performance on Spearman Dot Similarity}, is shown in Figure~\ref{fig:1} (f) where, the $paraphrase-multilingual-mpnet-base-v2$ and $LaBSE$ models showed higher values in Spearman Dot similarity, with noticeable decreases at lower dimensions. The $mpnet-base-all-nli-triplet$ model demonstrated moderate performance, with significant declines at lower dimensions. The lowest values were observed in $MARBERT$ and $bert-base-arabertv02$, indicating less consistency in capturing semantic similarity using Spearman Dot similarity.

Overall, the $paraphrase-multilingual-mpnet-base-v2$ model consistently outperformed others across various metrics, followed by $LaBSE$. The $mpnet-base-all-nli-triplet$ model showed moderate performance, while $MARBERT$ and $bert-base-arabertv02$ lagged, especially in capturing semantic similarity at lower dimensions.

\subsection{Comparison of Base Models Vs. Arabic Trained Matryoshka Models}

In this subsection, we aim to evaluate the performance of Matryoshka embedding models for Arabic by comparing the base models with their trained Matryoshka counterparts, we can discern how the nested embedding learning might enhance the models' ability to capture semantic similarity in the Arabic language. This analysis will be conducted across different evaluation metrics on 768 dimension, providing a comprehensive understanding of the improvements brought about by learning. To visually represent the impact of training Matryoshka Models, Figure~\ref{fig:1}, provide bar plots for each model comparing their performance against base models on two key metrics: Pearson Cosine Similarity and Spearman Cosine Similarity.

\begin{figure}[ht]
    \centering
    \begin{subfigure}[b]{0.18\textwidth}
        \centering
        \includegraphics[width=\textwidth]{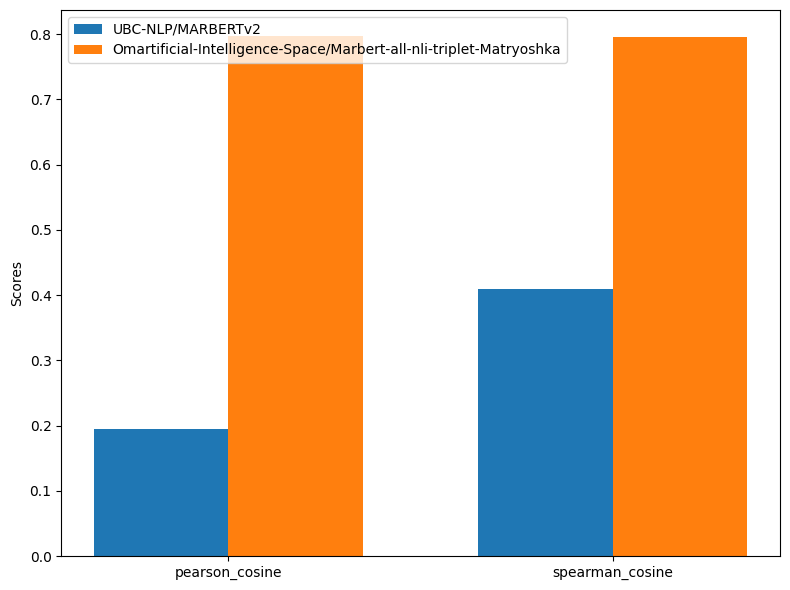}
        \caption{Marbert}
        \label{fig:marbert}
    \end{subfigure}
    \hfill
    \begin{subfigure}[b]{0.18\textwidth}
        \centering
        \includegraphics[width=\textwidth]{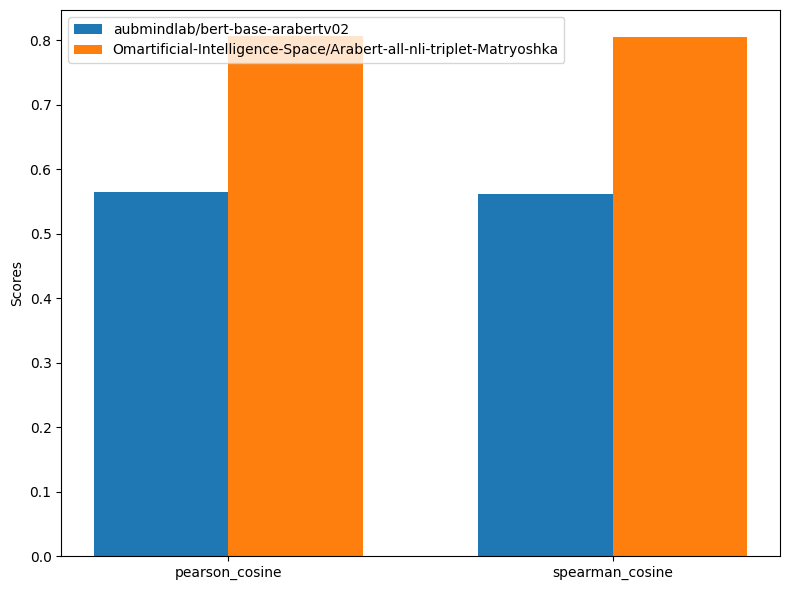}
        \caption{Arabert}
        \label{fig:arabert}
    \end{subfigure}
    \hfill
    \begin{subfigure}[b]{0.18\textwidth}
        \centering
        \includegraphics[width=\textwidth]{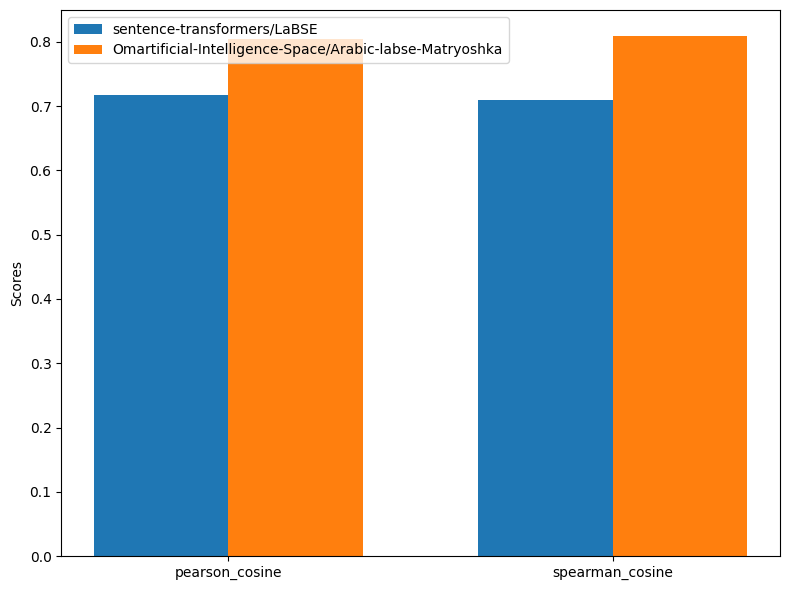}
        \caption{LaBSE}
        \label{fig:labse}
    \end{subfigure}
    \hfill
    \begin{subfigure}[b]{0.18\textwidth}
        \centering
        \includegraphics[width=\textwidth]{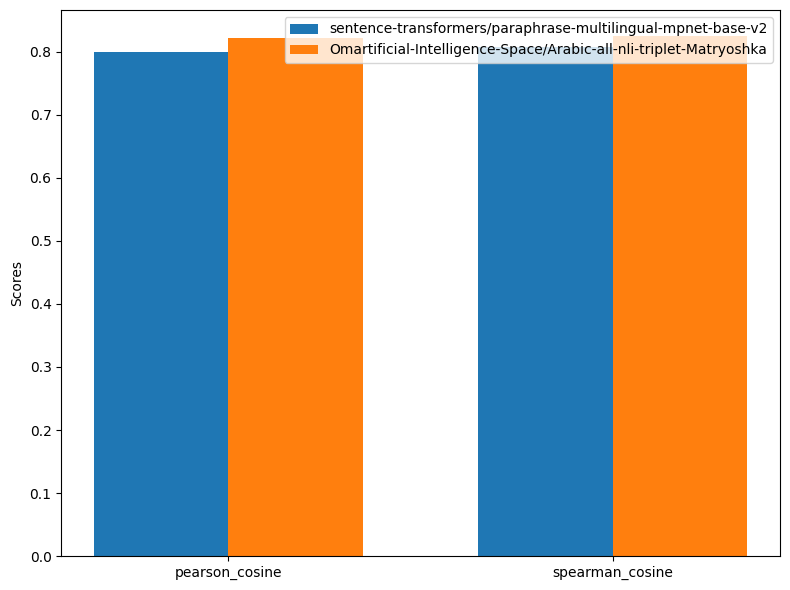}
        \caption{mpnet-base}
        \label{fig:multi}
    \end{subfigure}
    \hfill
    \begin{subfigure}[b]{0.18\textwidth}
        \centering
        \includegraphics[width=\textwidth]{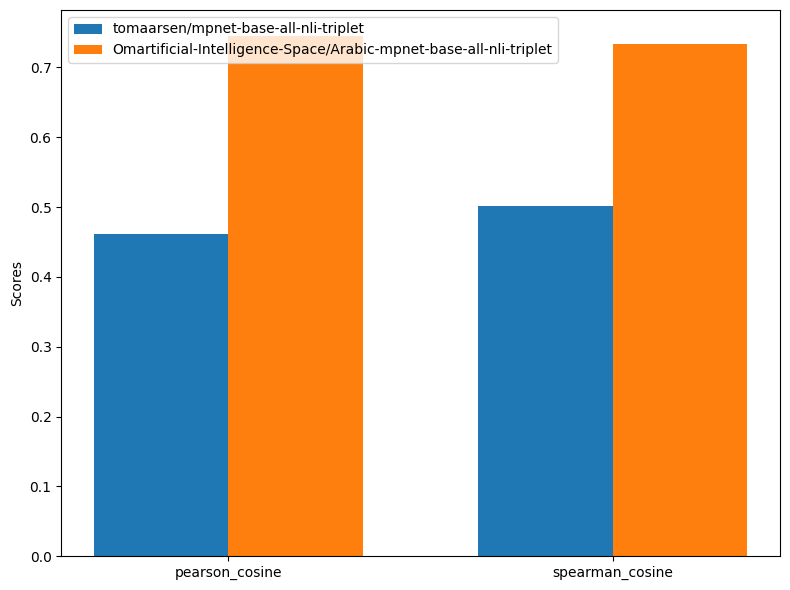}
        \caption{mpnet-En}
        \label{fig:mpnet}
    \end{subfigure}
    \caption{Comparison of Base Models Vs. Trained Matryoshka Models Across Various Metrics}
    \label{fig:2}
\end{figure}

As shown in Figure~\ref{fig:2} (a), the trained Matryoshka model  using MARBERT outperformed the base MARBERT model significantly in both Pearson and Spearman cosine similarity metrics. This substantial enhancement indicates that the Matryoshka version has a much-improved capability to capture semantic similarities, making it more suitable for nuanced language tasks in Arabic. Additionally, For the Arabert model shown in Figure~\ref{fig:2} (b), the trained Matryoshka version has also demonstrated significant improvements comparing to the base Arabert model. The enhanced scores across the cosine similarity metrics underscore the effectiveness of the Matryoshka model in refining the model's performance for Arabic text processing.

For the multilingual models shown in Figure~\ref{fig:2} (c) and (d), the trained Matryoshka model of LaBSE model showed improved performance compared to the base LaBSE model and also the trained multilingual mpnet base model also demonstrated superior performance over its base counterpart across both Pearson and Spearman cosine similarity metrics. This consistency suggests that the Matryoshka model enhances the model's adaptability to the Arabic language, making it more effective in capturing semantic similarities within this context.

Finally, the comparison for the English mpnet base model reveals that the Arabic Matryoshka version outperformed the base model  substantially. The improvements across Pearson and Spearman cosine metrics highlight the trained model's enhanced semantic understanding and accuracy for Arabic text.

\subsection{Analysis of Similarity Scores Predicted by Arabic Trained Matryoshka Models}

In this subsection, we analyze the similarity scores predicted by our Arabic trained Matryoshka models against the ground truth. The examples include cases of perfect similarity (score 1), no similarity (score 0), and moderate similarity (scores between 0 and 1). Table~\ref{tab:7}, \ref{tab:8} and \ref{tab:9} show the detailed examples and their corresponding analysis.

\begin{table}[ht]
\centering
\resizebox{\columnwidth}{!}{%
\begin{tabular}{|l|l|p{0.35\textwidth}|p{0.35\textwidth}|}
\hline
\textbf{Model} & \textbf{Score} & \textbf{Sentence1} & \textbf{Sentence2} \\ \hline
\multirow{2}{*}{Ground Truth} & \multirow{2}{*}{0.72} & \multirow{7}{*}{\begin{tabular}[c]{@{}c@{}} \RL{مجموعة من الرجال يلعبون كرة القدم} \\ (A group of men playing football) \end{tabular}} & \multirow{7}{*}{\begin{tabular}[c]{@{}c@{}} \RL{مجموعة من الأولاد يلعبون كرة القدم} \\ (A group of boys playing football) \end{tabular}} \\
 &  &  &  \\ \cline{1-2}
Arabic-mpnet-base-all-nli-triplet & 0.768 &  &  \\ \cline{1-2}
Arabic-all-nli-triplet-Matryoshka & 0.685 &  &  \\ \cline{1-2}
Arabert-all-nli-triplet-Matryoshka & 0.661 &  &  \\ \cline{1-2}
Arabic-labse-Matryoshka & 0.835 &  &  \\ \cline{1-2}
Marbert-all-nli-triplet-Matryoshka & 0.836 &  &  \\ \hline
\end{tabular}%
}
\caption{Comparison of model scores for the high similarity pair.}
\label{tab:7}
\end{table}

As shown in Table~\ref{tab:7}, with a ground truth score of 0.72, indicating moderate similarity, the models show varying degrees of accuracy. $Marbert-all-nli-triplet-Matryoshka$ and $Arabic-labse-Matryoshka$ predict slightly higher scores (0.836 and 0.835, respectively), closely aligning with the ground truth. This indicates the models' effectiveness in capturing moderate semantic similarities, particularly when sentences share significant contextual overlap.

\begin{table}[ht]
\centering
\resizebox{\columnwidth}{!}{%
\begin{tabular}{|l|l|p{0.35\textwidth}|p{0.35\textwidth}|}
\hline
\textbf{Model} & \textbf{Score} & \textbf{Sentence1} & \textbf{Sentence2} \\ \hline
\multirow{2}{*}{Ground Truth} & \multirow{2}{*}{0.1} & \multirow{7}{*}{\begin{tabular}[c]{@{}c@{}} \RL{رجل يعزف على الجيتار} \\ (A man playing the guitar) \end{tabular}} & \multirow{7}{*}{\begin{tabular}[c]{@{}c@{}} \RL{رجل يقود سيارة} \\ (A man driving a car) \end{tabular}} \\
 &  &  &  \\ \cline{1-2}
Arabic-mpnet-base-all-nli-triplet & 0.334 &  &  \\ \cline{1-2}
Arabic-all-nli-triplet-Matryoshka & 0.481 &  &  \\ \cline{1-2}
Arabert-all-nli-triplet-Matryoshka & 0.480 &  &  \\ \cline{1-2}
Arabic-labse-Matryoshka & 0.323 &  &  \\ \cline{1-2}
Marbert-all-nli-triplet-Matryoshka & 0.382 &  &  \\ \hline
\end{tabular}%
}
\caption{Comparison of model scores for the no similarity pair.}
\label{tab:8}
\end{table}

As shown in Table~\ref{tab:8}, for the ground truth score of 0.1, indicating no similarity, the Matryoshka models' scores are notably higher, ranging from 0.323 to 0.481. This suggests that while the models recognize some degree of unrelatedness, they still infer a minimal level of semantic similarity, possibly due to shared contextual elements like "\RL{رجل}" (man) in both sentences.

\begin{table}[ht]
\centering
\resizebox{\columnwidth}{!}{%
\begin{tabular}{|l|l|p{0.35\textwidth}|p{0.35\textwidth}|}
\hline
\textbf{Model} & \textbf{Score} & \textbf{Sentence1} & \textbf{Sentence2} \\ \hline
\multirow{2}{*}{Ground Truth} & \multirow{2}{*}{1} & \multirow{7}{*}{\begin{tabular}[c]{@{}c@{}} \RL{رجل يقوم بخدعة بالبطاقات} \\ (A man doing a card trick) \end{tabular}} & \multirow{7}{*}{\begin{tabular}[c]{@{}c@{}} \RL{رجل يقوم بخدعة ورق} \\ (A man performing a card trick) \end{tabular}} \\
 &  &  &  \\ \cline{1-2}
Arabic-mpnet-base-all-nli-triplet & 0.8 &  &  \\ \cline{1-2}
Arabic-all-nli-triplet-Matryoshka & 0.91 &  &  \\ \cline{1-2}
Arabert-all-nli-triplet-Matryoshka & 0.87 &  &  \\ \cline{1-2}
Arabic-labse-Matryoshka & 0.84 &  &  \\ \cline{1-2}
Marbert-all-nli-triplet-Matryoshka & 0.85 &  &  \\ \hline
\end{tabular}%
}
\caption{Comparison of model scores for the moderate similarity pair.}
\label{tab:9}
\end{table}

As shown in Table~\ref{tab:9}, the ground truth score is 1, indicating perfect similarity between the two sentences. All the Matryoshka models also predict high similarity scores, with $Arabic-all-nli-triplet-Matryoshka$ achieving the highest score of 0.906. This demonstrates the model's capability to capture near-perfect semantic similarity for sentences with minimal lexical variation.

\begin{figure}[ht]
\centering
\includegraphics[width=\textwidth]{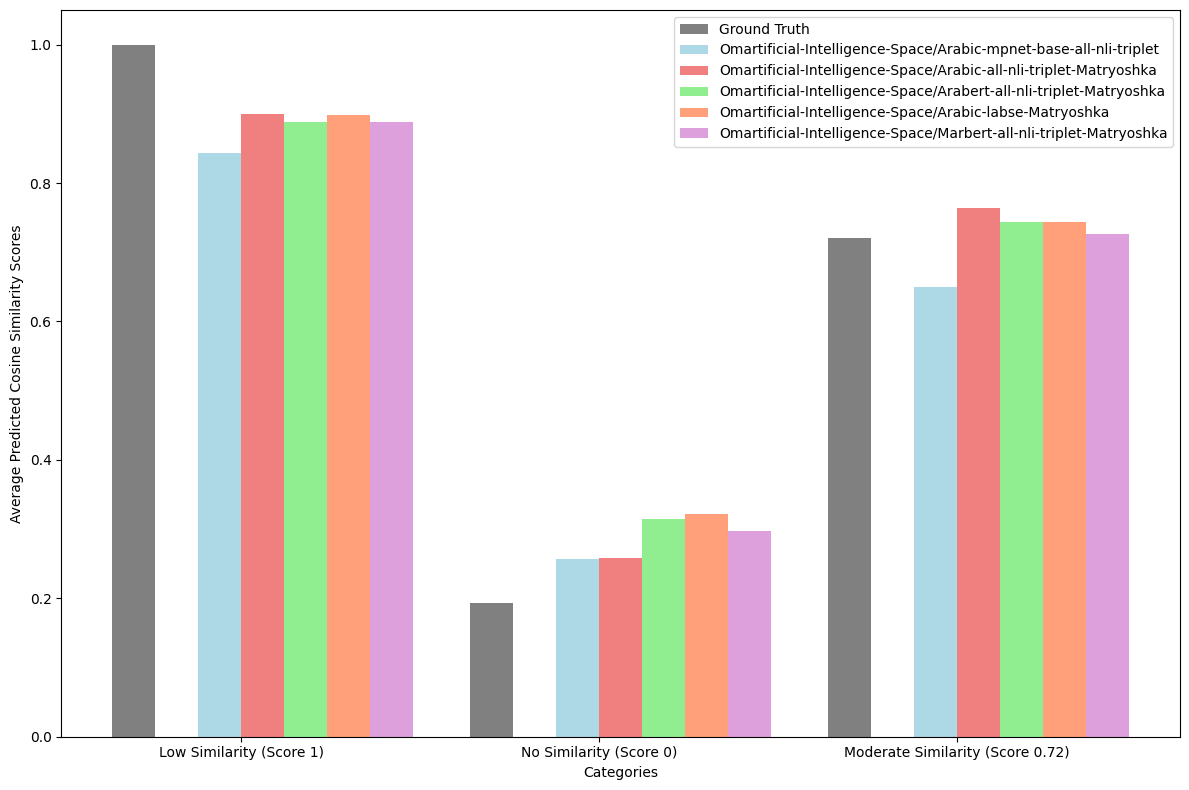}
\caption{Comparison of Average Predicted Cosine Similarity Scores for Different Similarity Categories}
\label{fig:3}
\end{figure}

To further analyze the performance of the trained Matryoshka models, we compare their average predicted similarity scores against the ground truth scores across three different similarity categories: low similarity, no similarity, and moderate similarity. The comparative results are illustrated in the Figure~\ref{fig:3}

The error analysis reveals that while the Arabic trained Matryoshka models generally perform well in identifying high and moderate similarity sentence pairs, they exhibit a tendency to overestimate similarity in the no similarity category. For the low similarity category, all models predicted scores close to the ground truth, demonstrating their effectiveness in capturing high similarity relationships. Specifically, the $Marbert-all-nli-triplet-Matryoshka$ and $Arabic-labse-Matryoshka$ models showed the highest accuracy in this category. However, in the no similarity category, the predicted scores were consistently higher than the ground truth, indicating false positives. 

This suggests a potential area for improvement, as the models tend to recognize some degree of similarity in dissimilar pairs. In the moderate similarity category, the models also performed well, although there was a slight tendency to overestimate similarity. Overall, these findings suggest that while the models are effective at capturing similarities, they require further refinement to accurately identify dissimilar pairs and reduce false positive rates. This enhancement can improve the robustness and accuracy of similarity models in multilingual contexts, especially for the Arabic language.

The consistent trend of improved performance across all trained Arabic Matryoshka models highlights the efficacy of the Nested learning process. The significant gains in both Pearson and Spearman cosine similarity metrics indicate that the trained models have a better grasp of the semantic nuances in the Arabic language. This enhancement makes these models more suitable for tasks requiring high precision in semantic similarity assessments, such as machine translation, information retrieval, and natural language understanding tasks specific to Arabic.

\begin{figure}[ht]
\centering
\includegraphics[width=\textwidth]{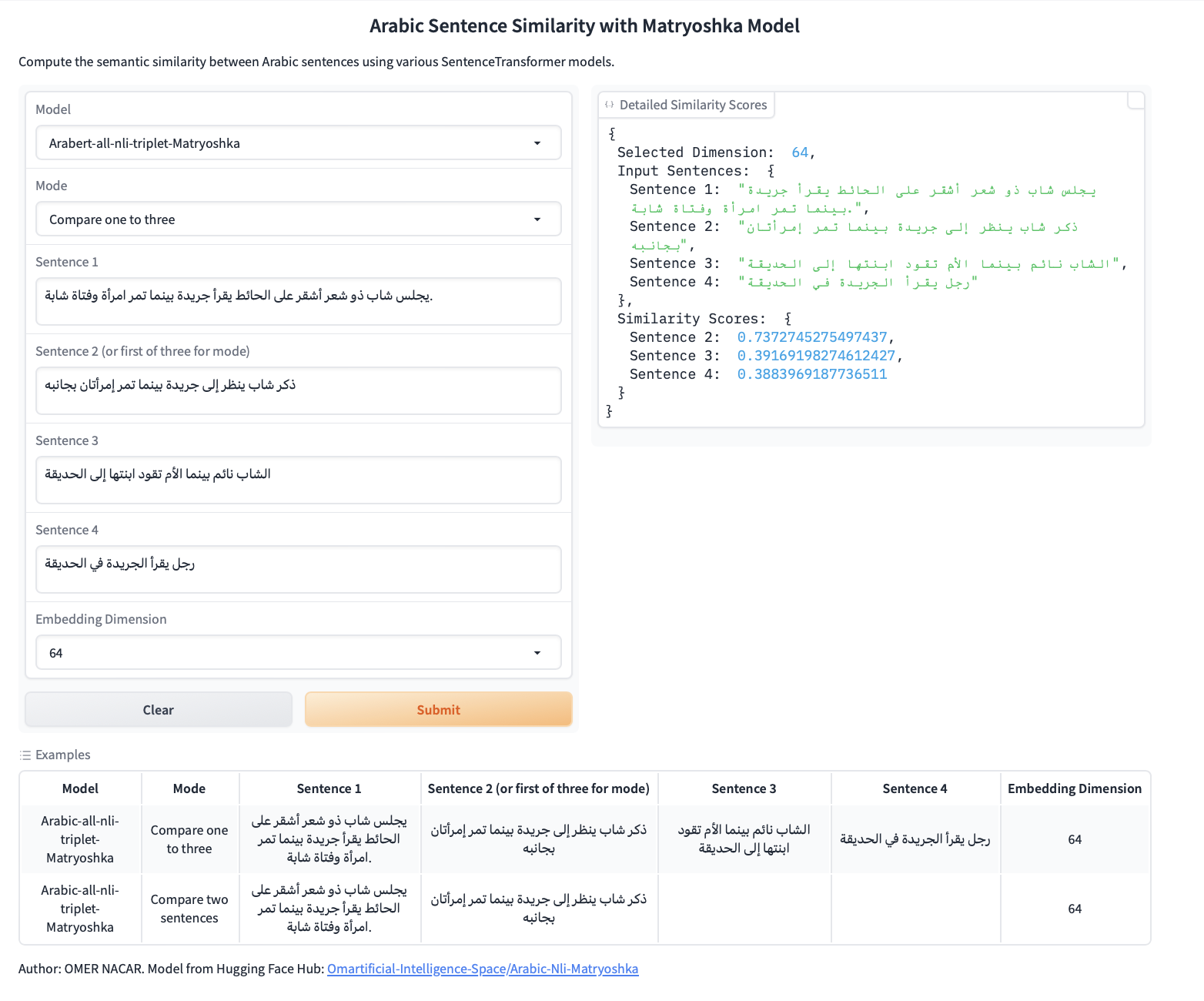}
\caption{User Interface of the Arabic Sentence Similarity Application }
\label{fig:4}
\end{figure}

\section{Arabic Sentence Similarity Application}

In addition to the comprehensive evaluation of the Arabic Matryoshka embedding models, we have developed an interactive $Gradio$ application that leverages these models to compute semantic similarity between Arabic sentences. This tool is designed to provide users with a practical interface to utilize the advanced capabilities of $SentenceTransformer$ models in real-world scenarios. Figure~\ref{fig:4} shows the user interface of the app depolyed in hugging-face\footnote{https://huggingface.co/spaces/Omartificial-Intelligence-Space/Arabic-Sentence-Similarity-Matryoshka-Models}. 

As shown in Figure~\ref{fig:4}, the app provides a variety of choices and modes along with the Arabic trained Matryoshka models combined together in the following Key Features;

\textbf{Model Selection}, users can choose from a variety of Arabic Matryoshka Embedding Models. This flexibility allows for comparison and selection of the most suitable model for specific needs.

\textbf{Flexible Comparison Modes}, the application supports two primary comparison modes. Users can either compare two sentences directly or compare one sentence against three others, offering a versatile approach to semantic similarity evaluation.

\textbf{Custom Embedding Dimensions}, to cater to various computational and accuracy requirements, the application allows users to select embedding dimensions from [768, 512, 256, 128, 64]. This feature ensures that the tool can be adapted to different performance and precision needs.

\textbf{Detailed Similarity Scores}, the application provides detailed similarity scores between sentences, enabling users to understand the nuances of the semantic relationships captured by the models.

By providing this application, we aim to bridge the gap between advanced model development and practical usability, making state-of-the-art sentence similarity computations accessible and user-friendly.

\section{Conclusion}
In this chapter, we conducted a thorough evaluation of multiple trained Arabic nested embedding models on the Arabic Semantic Textual Similarity Benchmark. Our analysis included models that are multilingual and those specifically trained for Arabic. Using the $EmbeddingSimilarityEvaluator$, we assessed their performance based on several metrics: Pearson and Spearman correlations for cosine similarity, Manhattan distance, Euclidean distance, and dot product similarity.

The results clearly demonstrate that tained arabic nested embedding models on Arabic data significantly improves their performance. The $paraphrase-multilingual-mpnet-base-v2$ model consistently outperformed others across most dimensions and metrics, highlighting its robustness in capturing semantic similarity. $LaBSE$ also showed stable performance, particularly in higher dimensions, making it a reliable choice for multilingual tasks. The $mpnet-base-all-nli-triplet$ model exhibited moderate performance, better than the $bert-base-arabertv02$ and $MARBERT$ models, which had the lowest values in most metrics. Our comparative analysis of the base models versus their Matryoshka counterparts further emphasizes the importance of language-specific training. The Matryoshka models showed marked improvements in capturing the semantic nuances of Arabic, as reflected in the significant increase in Pearson and Spearman correlations.

These findings underscore the necessity of adapting NLP models to specific languages to achieve optimal performance. The insights gained from this evaluation can guide future research and development efforts in creating more effective and nuanced language models for Arabic and other underrepresented languages.

\subsubsection*{Acknowledgments} The authors thank Prince Sultan University for their support.


\begin{thebibliography}{4}
\bibitem{LeCun}Y. LeCun, Y. Bengio, and G. Hinton. Deep learning. nature, 521(7553):436–444, 2015.
\bibitem{Nayak} P. Nayak. Understanding searches better than ever before. Google AI Blog, 2019. URL https: //blog.google/products/search/search-language-understanding-bert/.
\bibitem{Dean}J. Dean. Challenges in building large-scale information retrieval systems. In Keynote of the 2nd ACM International Conference on Web Search and Data Mining (WSDM), volume 10, 2009.
\bibitem{Sun}C. Sun, A. Shrivastava, S. Singh, and A. Gupta. Revisiting unreasonable effectiveness of data in deep learning era. In Proceedings of the IEEE international conference on computer vision, pages 843–852, 2017.
\bibitem{Kusupati}Kusupati, A., Bhatt, G., Rege, A., Wallingford, M., Sinha, A., Ramanujan, V., ... \& Farhadi, A. (2022). Matryoshka representation learning. Advances in Neural Information Processing Systems, 35, 30233-30249.
\bibitem{Nils}Nils Reimers and Iryna Gurevych. 2019. Sentence-bert: Sentence embeddings using siamese bert- networks. In Proceedings of the 2019 Conference on Empirical Methods in Natural Language Processing, pages 3980–3990. Association for Computational Linguistics.
\bibitem{Deng}J. Deng, W. Dong, R. Socher, L.-J. Li, K. Li, and L. Fei-Fei. Imagenet: A large-scale hierarchical image database. In 2009 IEEE conference on computer vision and pattern recognition, pages 248–255. Ieee, 2009.
\bibitem{He}K. He, X. Zhang, S. Ren, and J. Sun. Deep residual learning for image recognition. In Proceedings of the IEEE conference on computer vision and pattern recognition, pages 770– 778, 2016.
\bibitem{Gutmann}M. Gutmann and A. Hyvärinen. Noise-contrastive estimation: A new estimation principle for unnormalized statistical models. In Proceedings of the thirteenth international conference on artificial intelligence and statistics, pages 297–304. JMLR Workshop and Conference Proceedings, 2010.
\bibitem{Harris}M. G. Harris and C. D. Giachritsis. Coarse-grained information dominates fine-grained information in judgments of time-to-contact from retinal flow. Vision research, 40(6):601–611, 2000.
\bibitem{Waldburger}C. Waldburger. As search needs evolve, microsoft makes ai tools for better search available to researchers and developers. Microsoft AI Blog, 2019. URL https://blogs.microsoft. com/ai/bing-vector-search/.
\bibitem{Malkov}Y. A. Malkov and D. A. Yashunin. Efficient and robust approximate nearest neighbor search using hierarchical navigable small world graphs. IEEE transactions on pattern analysis and machine intelligence, 42(4):824–836, 2018.
\bibitem{Cai} H. Cai, C. Gan, T. Wang, Z. Zhang, and S. Han. Once-for-all: Train one network and specialize it for efficient deployment. arXiv preprint arXiv:1908.09791, 2019.
\bibitem{Li}M. Wallingford, H. Li, A. Achille, A. Ravichandran, C. Fowlkes, R. Bhotika, and S. Soatto. Task adaptive parameter sharing for multi-task learning. arXiv preprint arXiv:2203.16708, 2022.
\bibitem{Yu}  J. Yu, L. Yang, N. Xu, J. Yang, and T. Huang. Slimmable neural networks. arXiv preprint arXiv:1812.08928, 2018.
\bibitem{Rippel}O. Rippel, M. Gelbart, and R. Adams. Learning ordered representations with nested dropout. In International Conference on Machine Learning, pages 1746–1754. PMLR, 2014.
\bibitem{Bowman}Bowman, S. R., Angeli, G., Potts, C., \& Manning, C. D. (2015). A large annotated corpus for learning natural language inference. arXiv preprint arXiv:1508.05326.
\bibitem{Kim}Kim, S., Kang, I., \& Kwak, N. (2019, July). Semantic sentence matching with densely-connected recurrent and co-attentive information. In Proceedings of the AAAI conference on artificial intelligence (Vol. 33, No. 01, pp. 6586-6593).
\bibitem{MacCartney}MacCartney, B., \& Manning, C. D. (2008, August). Modeling semantic containment and exclusion in natural language inference. In Proceedings of the 22nd International Conference on Computational Linguistics (Coling 2008) (pp. 521-528).
\bibitem{Yang}Yang, Y., Yuan, S., Cer, D., Kong, S. Y., Constant, N., Pilar, P., ... \& Kurzweil, R. (2018). Learning semantic textual similarity from conversations. arXiv preprint arXiv:1804.07754.
\bibitem{Kudo}Kudo, T., \& Richardson, J. (2018). Sentencepiece: A simple and language independent subword tokenizer and detokenizer for neural text processing. arXiv preprint arXiv:1808.06226.
\bibitem{Klein}Klein, G., Kim, Y., Deng, Y., Senellart, J., \& Rush, A. M. (2017). Opennmt: Open-source toolkit for neural machine translation. arXiv preprint arXiv:1701.02810.
\bibitem{Henderson} Henderson, M., Al-Rfou, R., Strope, B., Sung, Y. H., Lukács, L., Guo, R., ... \& Kurzweil, R. (2017). Efficient natural language response suggestion for smart reply. arXiv preprint arXiv:1705.00652.

\end{thebibliography}
\end{document}